\documentclass{article}
\usepackage{spconf,amsmath,graphicx,hyperref}
\usepackage{booktabs}
\usepackage{placeins}
\newcommand{\best}[1]{\textbf{#1}}
\newcommand{\pfgs}{\textsc{PFGS}}

\title{PHONEME-GUIDED TTS AUGMENTATION FOR ASR: A UNIFIED PIPELINE AND MULTILINGUAL EVALUATION}

\name{Zhen Wang$^{1}$, Tianrui Wu$^{1}$, Rongqi Han$^{1}$, Hao Wu$^{1}$, Wei Liang$^{2}$, Wei Xu$^{1,3}$}
\address{$^{1}$Shanghai Qi Zhi Institute, Shanghai, China\\
$^{2}$Megatronix (Beijing) Technology Co., Ltd.\\
$^{3}$Tsinghua University, Beijing, China}

\begin{document}
\ninept
\maketitle

\begin{abstract}
Synthetic speech can provide additional supervision for automatic speech recognition (ASR), but constructing useful synthetic training data requires choosing both what to synthesize and how to synthesize it. We present a phoneme-guided text-to-speech (TTS) augmentation pipeline for ASR that connects multilingual speech generation with candidate-text selection and reference-speech quality control. Within this pipeline, we propose phoneme-frequency-guided selection (\pfgs{}), which uses phoneme frequencies from real ASR training transcripts to prioritize candidate texts containing common phonetic content. Experiments with separate monolingual ASR systems cover four languages and 13 test sets. With random text selection, the pipeline improves recognition on 11 test sets at one or more synthesis ratios. \pfgs{} further outperforms random selection on nine test sets, with relative word error rate (WER) reductions of up to 19.3\%. An ablation with fixed target texts and synthesis counts further shows the benefit of reference-speech filtering. These results support using real-data phoneme statistics to guide the construction of effective synthetic supervision for ASR.
\end{abstract}


\section{Introduction}
\label{sec:introduction}

Text-to-speech (TTS) augmentation converts text resources into paired speech and transcripts, providing additional supervision when transcribed speech or target-domain data are limited~\cite{bartelds2023making,huang2023textgen,yang2025versatile}. Recent zero-shot TTS models, including F5-TTS and CosyVoice, further improve the naturalness, speaker diversity, and controllability of synthesized speech~\cite{chen2025f5tts,du2024cosyvoice}.

The downstream value of synthetic speech nevertheless depends on what is synthesized, how it is synthesized, and how much is added to ASR training. Previous studies have selected unspoken text using language models, generated new text with neural models, or learned to select useful synthesized utterances~\cite{chen2020contrastive,huang2023textgen,liu2023selection}. Hard-Synth combines LLM-based text rewriting with reference-prompt selection guided by ASR difficulty~\cite{yu2024hardsynth}. Other studies have examined synthetic-data quantity and diversity~\cite{yang2025versatile} or improved synthetic-speech quality through filtering~\cite{perrin2025optimized}. Phonetic content has also guided sentence selection through iterative matching of the combined real and synthetic diphone distribution to a target distribution~\cite{ogun2025exhaustive}.

These studies motivate treating TTS augmentation as a synthetic-corpus construction problem encompassing linguistic content, acoustic conditions, and synthesis scale. Within this perspective, phoneme information can connect speech generation with candidate-text selection, serving both as a TTS input representation and as a selection criterion. A practical question is how a pipeline built around these two roles performs across languages and evaluation domains as synthesis scale, text selection, and reference quality vary.

We develop an ASR-oriented synthetic-data construction pipeline that uses phoneme information in both speech generation and candidate-text selection. The pipeline combines a shared multilingual phoneme-based TTS model with reference-speech quality control. Phoneme frequencies offer a simple basis for deciding which phonetic content to prioritize for synthesis. We explore whether emphasizing phonemes that occur frequently in real ASR training transcripts can make synthetic supervision more effective. To this end, we introduce phoneme-frequency-guided selection (\pfgs{}), a lightweight method that uses these frequencies to rank candidate texts before synthesis. 

We evaluate the pipeline using separate monolingual ASR systems for four languages across 13 test sets. We first assess the pipeline’s effectiveness using random text selection across synthesis scales. We then examine whether \pfgs{} provides further gains over random text selection and text selection favoring rare phonemes. Finally, we evaluate the contribution of reference-speech filtering with target texts and synthesis counts held fixed.

\section{Method}
\label{sec:method}

\subsection{Overall Pipeline and Data Objects}
\label{ssec:pipeline}

\begin{figure*}[t]
    \centering
    \includegraphics[width=\textwidth]{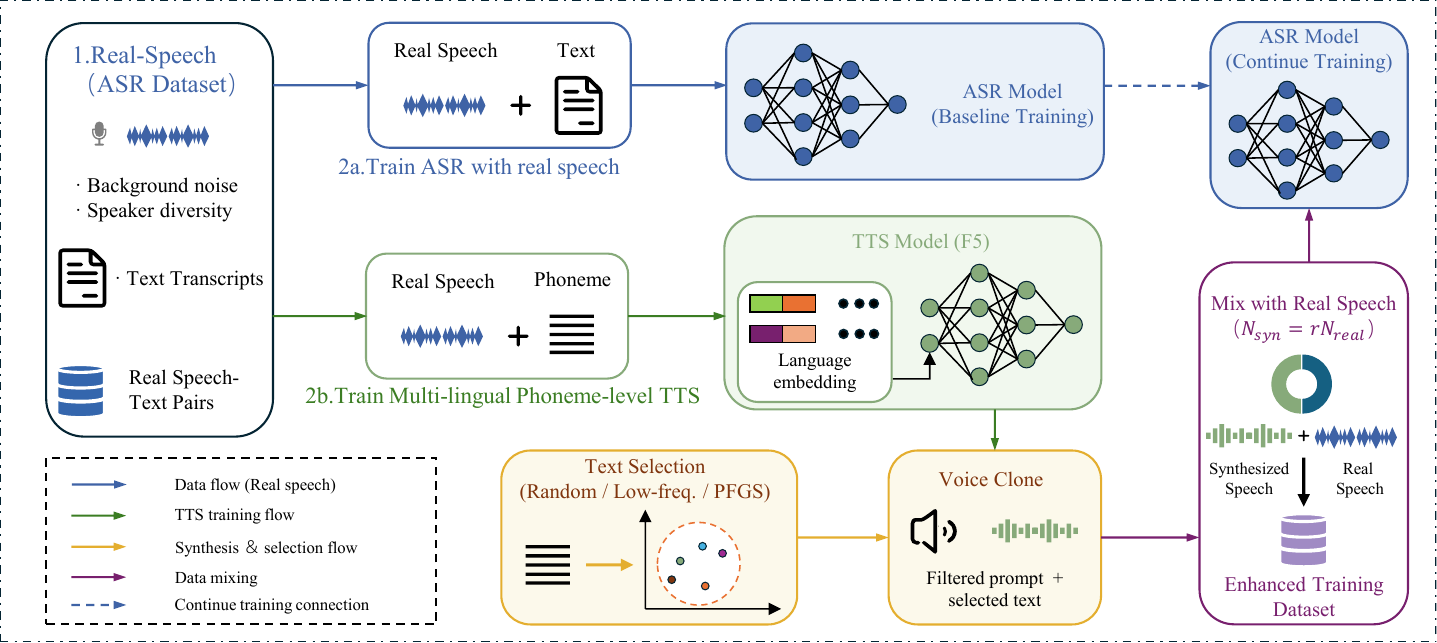}
    \caption{The phoneme-guided TTS-to-ASR augmentation pipeline. Candidate texts are selected before synthesis, and quality-filtered reference prompts condition the shared multilingual TTS model.}
    \label{fig:pipeline}
\end{figure*}

The ASR-oriented TTS augmentation pipeline comprises five stages. These cover data preparation, phoneme-based TTS training, reference-speech quality control and task construction, synthetic-corpus control, and ASR training and evaluation. Figure~\ref{fig:pipeline} summarizes the complete procedure.

We use three data objects:
\begin{equation}
\begin{aligned}
D_{\mathrm{TTS}} &= \{(a_j,u_j,l_j)\},\\
D_{\mathrm{ASR}}^{\mathrm{real}} &= \{(x_i,y_i)\},\\
T_{\mathrm{cand}}^{(l)}
&= \operatorname{ExcludeTest}\!\big(
\operatorname{Clean}\!\left(T_{\mathrm{source}}^{(l)}\right), T_{\mathrm{test}}^{(l)}\big).
\end{aligned}
\label{eq:data-objects}
\end{equation}
Here, $(a_j,u_j,l_j)$ denotes natural speech, its transcript, and language identifier, while $(x_i,y_i)$ is a real ASR utterance--label pair. $D_{\mathrm{TTS}}$ trains the multilingual TTS model, and $D_{\mathrm{ASR}}^{\mathrm{real}}$ provides ASR supervision and phoneme statistics. Candidate texts are drawn from the real ASR training transcripts and other available text sources. For each language, all evaluated selection strategies operate on the same candidate pool. $\operatorname{Clean}(\cdot)$ applies language identification, normalization, length and character checks, and deduplication, while $\operatorname{ExcludeTest}(\cdot)$ removes test-transcript overlap. \pfgs{} ranks $T_{\mathrm{cand}}^{(l)}$ using the real-data phoneme statistics and selects $N_{\mathrm{syn}}$ targets. Each target is synthesized with a filtered reference prompt to yield an ASR pair $(x_{\mathrm{syn}},t)$. The resulting synthetic dataset $D_{\mathrm{syn}}$ is combined with $D_{\mathrm{ASR}}^{\mathrm{real}}$ to form the augmented training set $D_{\mathrm{ASR}}^{\mathrm{aug}}$.

\subsection{Multilingual Natural Speech and Phoneme Representation}
\label{ssec:phoneme-tts}

We train a shared phoneme-based TTS model from scratch with the F5-TTS architecture~\cite{chen2025f5tts} on naturally recorded Arabic, French, Italian, and Portuguese speech. The corpus preserves speaker, channel, and background variation; only samples with unreliable alignment, abnormal duration, or poor text quality are excluded.

Input text first undergoes language-specific normalization and grapheme-to-phoneme (G2P) conversion. Arabic, French, and Italian use eSpeak, whereas Portuguese uses gruut.
To distinguish languages within the shared model, we add a 32-dimensional learnable language embedding to the original F5-TTS input layer. For language $l$, the embedding is expanded over time. It is then concatenated with the noisy mel state $x_t$, reference-speech conditioning mel $c_{\mathrm{ref}}$, and 512-dimensional phoneme text embedding $e_{\mathrm{text}}$:
\begin{equation}
h_0 = \operatorname{Proj}\!\left(x_t \mathbin{\Vert} c_{\mathrm{ref}} \mathbin{\Vert} e_{\mathrm{text}} \mathbin{\Vert} \operatorname{Expand}\!\left(e_{\mathrm{lang}}(l)\right)\right)
\label{eq:language-embedding}
\end{equation}
Here, $\Vert$ denotes feature-wise concatenation. A linear layer projects the resulting 744-dimensional representation to 1024 dimensions before the diffusion transformer backbone. The language embedding identifies the target language, while the phoneme sequence provides a common pronunciation representation across writing systems.

Reference prompts are drawn from a pool of natural speech. We sample reference prompts from the language-specific pools to expose the ASR model to a range of speaker characteristics through synthetic speech. We retain utterances lasting 3--12 seconds, containing at least three words, and having transcript-length-to-duration ratios of 3--25 characters per second.  A pretrained Whisper large-v3 model, run with Faster-Whisper, then checks consistency between each reference recording and its original transcript~\cite{radford2023whisper}.
For each target text $t$, we select reference audio $a_{\mathrm{ref}}$ and its transcript $u_{\mathrm{ref}}$ from the corresponding language-specific prompt pool. The resulting synthesis task is
\begin{equation}
q = (a_{\mathrm{ref}},u_{\mathrm{ref}},t,l).
\label{eq:synthesis-task}
\end{equation}
Within each language, the synthesis-scale and text-selection experiments share the TTS checkpoint, prompt-pool construction procedure, reference-speech sampling rules, and inference parameters.

\subsection{Phoneme-Frequency-Guided Selection}
\label{ssec:pfgs}

\pfgs{} estimates a phoneme prior from the real ASR training labels. For each phoneme $p$,
\begin{equation}
\begin{aligned}
c(p)&=\operatorname{count}\!\left(p,\{\operatorname{G2P}(y_i)\}\right),\\
P_{\mathrm{valid}}&=\{p\in P\setminus P_{\mathrm{sp}}\mid c(p)\ge\tau\},\\
\pi(p)&=\frac{c(p)}{\sum_{q\in P_{\mathrm{valid}}}c(q)}.
\end{aligned}
\label{eq:phone-prior}
\end{equation}
Here, $P$ is the phoneme inventory, and $P_{\mathrm{sp}}$ contains special symbols. We set $\tau=200$ to exclude extremely low-count units. For candidate text $t$ with phoneme sequence $\phi(t)=\operatorname{G2P}(t)$, \pfgs{} computes
\begin{equation}
s_{\mathrm{freq}}(t)=\frac{1}{|\phi(t)|}\sum_{p\in\phi(t)}\pi(p),
\label{eq:pfgs-score}
\end{equation}
where $\pi(p)=0$ outside $P_{\mathrm{valid}}$. We rank candidates by decreasing score and synthesize the top $N_{\mathrm{syn}}$ sentences. By assigning greater weight to phonemes that occur more frequently in the real ASR training transcripts, \pfgs{} favors candidate texts containing a higher proportion of these phonemes.

\section{Experimental Setup}
\label{sec:experiments}

\subsection{Data}
\label{ssec:data}

TTS generation and downstream ASR augmentation cover Arabic, French, Italian, and Portuguese. Table~\ref{tab:training-scale} reports the training scale for each language.

\begin{table}[t]
\centering
\caption{Training-data scale for the four target languages.}
\label{tab:training-scale}
\fontsize{9}{11}\selectfont
\setlength{\tabcolsep}{2pt}
\renewcommand{\arraystretch}{1.0}

\begin{tabular*}{\columnwidth}
{@{\extracolsep{\fill}}lrrrr@{}}
\toprule
Language & TTS utt. & TTS h & ASR utt. & ASR h \\
\midrule
Arabic     & 1,259,489 & 1,899.67 & 350,000 & 563.438 \\
French     &   826,261 & 1,752.22 & 300,000 & 431.247 \\
Italian    &   322,054 &   653.25 & 172,469 & 254.102 \\
Portuguese &   559,205 &   626.62 &  22,348 &  25.667 \\
\midrule
Total      & 2,967,009 & 4,931.76 & 844,817 & 1,274.454 \\
\bottomrule
\end{tabular*}
\end{table}

Candidate pools combine the corresponding real ASR training transcripts with available transcripts from the following corpora. Arabic uses an internal, non-public corpus. French uses Common Voice, MLS, M-AILABS, and FLEURS. Italian uses Common Voice, MLS, M-AILABS, and VoxPopuli, while Portuguese uses CORAA, NURC-SP, MLS, and Common Voice~\cite{ardila2020commonvoice,conneau2022fleurs,pratap2020mls,solak2019mailabs,wang2021voxpopuli,candido2023coraa,rodrigues2024nurc}. Evaluation uses the applicable public Common Voice, FLEURS, MLS, SADA, and MASC test sets~\cite{alharbi2024sada,alfetyani2023masc}. Training, development, and test partitions are disjoint, and test transcripts are removed before candidate selection.

\begin{table*}[!t]
\centering
\caption{WER (\%) for random TTS augmentation at different synthesis ratios. CV denotes Common Voice; bold indicates the lowest WER within each language and test set.}
\label{tab:ratio}
\fontsize{9}{11}\selectfont
\setlength{\tabcolsep}{3pt}
\renewcommand{\arraystretch}{1.0}

\begin{tabular*}{\textwidth}
{@{\extracolsep{\fill}}llccccc@{}}
\toprule
Language & Condition & CV & FLEURS & MLS & SADA & MASC \\
\midrule
Arabic & Real-only (0\%) & \best{24.79} & 22.18 & -- & 62.48 & 52.82 \\
 & TTS 10\%  & 25.17 & 19.35 & -- & 59.98 & 52.92 \\
 & TTS 30\%  & 25.20 & 18.90 & -- & 59.19 & 52.78 \\
 & TTS 60\%  & 25.41 & \best{18.54} & -- & 58.80 & \best{52.59} \\
 & TTS 100\% & 25.11 & 18.59 & -- & \best{58.63} & 52.65 \\
\midrule
French & Real-only (0\%) & 11.50 & 23.07 & 25.24 & -- & -- \\
 & TTS 10\%  & 11.53 & 22.90 & 26.14 & -- & -- \\
 & TTS 30\%  & 11.34 & 21.48 & 26.48 & -- & -- \\
 & TTS 60\%  & 11.28 & 22.03 & \best{24.57} & -- & -- \\
 & TTS 100\% & \best{11.11} & \best{20.88} & 25.59 & -- & -- \\
\midrule
Italian & Real-only (0\%) & 11.05 & 11.41 & \best{33.92} & -- & -- \\
 & TTS 10\%  & 11.11 & 11.51 & 35.28 & -- & -- \\
 & TTS 30\%  & 11.05 & 11.28 & 35.69 & -- & -- \\
 & TTS 60\%  & 10.96 & 11.42 & 34.43 & -- & -- \\
 & TTS 100\% & \best{10.91} & \best{11.08} & 34.58 & -- & -- \\
\midrule
Portuguese & Real-only (0\%) & 49.12 & 56.10 & 75.44 & -- & -- \\
 & TTS 10\%  & 47.07 & 56.96 & 75.75 & -- & -- \\
 & TTS 30\%  & 45.78 & 55.25 & 74.16 & -- & -- \\
 & TTS 60\%  & 43.30 & 53.46 & 74.56 & -- & -- \\
 & TTS 100\% & \best{40.09} & \best{48.74} & \best{69.12} & -- & -- \\
\bottomrule
\end{tabular*}
\end{table*}

\subsection{Evaluation Setup}
\label{ssec:evaluation}

Each language uses an independently trained WeNet-based hybrid CTC/attention Conformer~\cite{yao2021wenet,gulati2020conformer}. For computational efficiency, real-only and TTS-augmented branches continue from the same language-specific checkpoint on $D_{\mathrm{ASR}}^{\mathrm{real}}$ and $D_{\mathrm{ASR}}^{\mathrm{aug}}$, respectively. They share the architecture, optimization settings, continuation interval, and decoding configuration. The intervals are epochs 70--100 for French, 80--140 for Arabic, and 70--180 for Italian and Portuguese. We evaluate ASR performance using word error rate (WER). All primary results use attention rescoring with a beam size of 10. For Arabic, diacritics are removed from the ASR training transcripts and from references and hypotheses before WER computation.

\subsection{Experiment 1: Synthesis-Scale Sweep}

We conduct a synthesis-scale sweep by randomly sampling candidate texts according to
\begin{equation}
{N_{\mathrm{syn}}}=r{N_{\mathrm{real}}},\qquad
r\in\{0\%,10\%,30\%,60\%,100\%\}.
\label{eq:synthesis-ratio}
\end{equation}
Here, $r=0\%$ denotes matched real-only continuation. $N_{\mathrm{syn}}$ and $N_{\mathrm{real}}$ denote the numbers of added synthetic utterances and real ASR training utterances, respectively. For $r>0$, target texts are sampled randomly from each language's candidate set. We compare the resulting ASR performance under the same language-specific training protocol.

\subsection{Experiment 2: PFGS and Control Strategies}
\label{ssec:selection-setup}
Following the evaluation of the pipeline with random text selection, we examine whether \pfgs{} further improves ASR performance. At a nominal 60\% utterance ratio, we compare \pfgs{} with random text selection and a low-frequency control that favors rare phonemes, using real-only training as the baseline. Random selection samples candidates uniformly, whereas the low-frequency control ranks candidate texts by the number of occurrences of rare phonemes. \pfgs{} follows the scoring procedure in Section~\ref{ssec:pfgs}. The low-frequency control was evaluated for Arabic, Italian, and Portuguese; the French experiments included real-only training, random selection, and \pfgs{}. All synthesis conditions share the same TTS checkpoint, prompt pool, and ASR training protocol.

\begin{table*}[!t]
\centering
\caption{WER (\%) by text-selection strategy. Percentages in condition labels give realized utterance/duration ratios relative to real training data; values beside \pfgs{} WERs give relative WER reductions from Random-60. CV denotes Common Voice; bold marks the lowest WER.}
\label{tab:pfgs}
\fontsize{9}{11}\selectfont
\setlength{\tabcolsep}{3pt}
\renewcommand{\arraystretch}{1.0}

\begin{tabular*}{\textwidth}
{@{\extracolsep{\fill}}llccccc@{}}
\toprule
Language & Condition & CV & FLEURS & MLS & SADA & MASC \\
\midrule
Arabic & Real-only & 24.79 & 22.18 & -- & 62.48 & 52.82 \\
 & Random (60.00\%/55.43\%) & 25.41 & 18.54 & -- & \best{58.80} & 52.59 \\
 & Low-frequency (57.14\%/74.66\%) & 25.25 & \best{17.42} & -- & 58.91 & 52.61 \\
 & \pfgs{} (57.14\%/35.12\%) & \shortstack{\best{24.73}(+2.7\%)} & \shortstack{19.49($-5.1\%$)} & -- & \shortstack{60.62($-3.1\%$)} & \shortstack{\best{52.29}(+0.6\%)} \\
\midrule
French & Real-only & 11.50 & 23.07 & 25.24 & -- & -- \\
 & Random (60.00\%/45.20\%) & 11.28 & 22.03 & 24.57 & -- & -- \\
 & \pfgs{} (61.46\%/60.27\%) & \shortstack{\best{11.11}(+1.5\%)} & \shortstack{\best{18.36}(+16.7\%)} & \shortstack{\best{19.95}(+18.8\%)} & -- & -- \\
\midrule
Italian & Real-only & 11.05 & 11.41 & 33.92 & -- & -- \\
 & Random (60.00\%/48.29\%) & \best{10.96} & 11.42 & 34.43 & -- & -- \\
 & Low-frequency (57.21\%/55.36\%) & 11.86 & 11.60 & 35.91 & -- & -- \\
 & \pfgs{} (56.28\%/58.44\%) & \shortstack{11.16($-1.8\%$)} & \shortstack{\best{10.92}(+4.4\%)} & \shortstack{\best{31.05}(+9.8\%)} & -- & -- \\
\midrule
Portuguese  & Real-only & 49.12 & 56.10 & 75.44 & -- & -- \\
 & Random (60.00\%/44.51\%) & \best{43.30} & 53.46 & 74.56 & -- & -- \\
 & Low-frequency (62.47\%/57.24\%) & 47.82 & 54.21 & 74.76 & -- & -- \\
 & \pfgs{} (64.23\%/59.94\%) & \shortstack{43.76($-1.1\%$)} & \shortstack{\best{49.70}(+7.0\%)} & \shortstack{\best{60.15}(+19.3\%)} & -- & -- \\
\bottomrule
\end{tabular*}
\end{table*}

\subsection{Experiment 3: Ablation of Reference-Speech Filtering}
\label{ssec:prompt-ablation-setup}

The reference-speech ablation compares filtered and unfiltered prompts on French and Italian Common Voice. The two conditions use identical \pfgs{} target-text lists and synthesis counts, differing only in reference-speech filtering.

\section{Results and Analysis}
\label{sec:results}

\subsection{Effect of Synthesis Scale under Random Text Selection}
\label{ssec:ratio-results}
 
Table~\ref{tab:ratio} summarizes the random-ratio sweep across 13 test sets. At least one augmented condition outperformed matched real-only continuation on 11 sets. The best condition used a 100\% ratio on eight sets, a 60\% ratio on three, and real-only training on two. Portuguese showed the clearest scale-dependent gains. At 100\%, WER decreased by 9.03, 7.36, and 6.32 absolute points on Common Voice, FLEURS, and MLS, respectively. Arabic Common Voice and Italian MLS did not improve at any augmentation ratio.   Overall, these results support the effectiveness of the proposed phoneme-based TTS augmentation pipeline with random text selection across multiple languages and test domains.

\subsection{Comparison of PFGS and Control Strategies}
\label{ssec:pfgs-results}

Under the nominal 60\% synthesis budget, Table~\ref{tab:pfgs} compares \pfgs{} with random and low-frequency text selection. \pfgs{} outperformed random selection on nine of the 13 test sets, with relative WER reductions of 0.6--19.3\% among these improvements. It also outperformed real-only training on 12 sets and low-frequency selection in eight of the ten available comparisons. The largest gains over random selection occurred on French FLEURS and MLS, Italian MLS, and Portuguese FLEURS and MLS. Arabic showed greater domain variation. \pfgs{} performed best on Common Voice and MASC, whereas low-frequency and random selection performed best on FLEURS and SADA, respectively. These comparisons show that prioritizing frequent phonemes can be an effective text-selection strategy for TTS augmentation. \pfgs{} outperformed low-frequency selection in most evaluated settings, indicating that additional supervision for common phonetic content remains useful even when those phonemes are already well represented in the real training data.

\subsection{Ablation of Reference-Speech Filtering}
\label{ssec:prompt-ablation-results}

We evaluated reference-speech filtering while holding target texts, synthesis counts, and training protocols fixed. Under attention rescoring on Common Voice, filtering reduced WER from 11.45 to 11.16 for Italian. It also reduced WER from 11.70 to 11.11 for French. These changes correspond to absolute reductions of 0.29 and 0.59 points, respectively. Filtering therefore reduced WER in both evaluated settings.

\FloatBarrier
\section{Conclusion}
\label{sec:conclusion}
We presented a phoneme-guided TTS-to-ASR augmentation pipeline that connects speech generation and candidate-text selection through phoneme information. Experiments across four languages demonstrated the effectiveness of the pipeline and the benefits of \pfgs{} and reference-speech filtering. These findings support a practical approach to ASR augmentation in which phoneme statistics from real training transcripts guide the construction of synthetic supervision, with frequently occurring phonetic content serving as a productive target for augmentation.

\bibliographystyle{IEEEbib}
\bibliography{paper_refs}

\end{document}